# Deep Encoder-Decoder Neural Network for Fingerprint Image Denoising and Inpainting

**Weiya Fan**

fanweiya@cqnu.edu.cn

**Chongqing Normal University,China**

**ABSTRACT**: Fingerprint image denoising is a very important step in fingerprint identification. to improve the denoising effect of fingerprint image,we have designs a fingerprint denoising algorithm based on deep encoder-decoder network,which encoder subnet to learn the fingerprint features of noisy images.the decoder subnet reconstructs the original fingerprint image based on the features to achieve denoising, while using the dilated convolution in the network to increase the receptor field without increasing the complexity and improve the network inference speed. In addition, feature fusion at different levels of the network is achieved through the introduction of residual learning, which further restores the detailed features of the fingerprint and improves the denoising effect. Finally, the experimental results show that the algorithm enables better recovery of edge, line and curve features in fingerprint images, with better visual effects and higher peak signal-to-noise ratio (PSNR) compared to other methods.

## 0 INTRODUCTION

With the development of various electronic devices such as mobile phones and tablets nowadays, automatic fingerprint identification technology as an advanced secure identification and access system has become popular among these electronic devices. Fingerprint recognition system is based on the detailed features of the fingerprint, how the fingerprint image is processed and whether the detailed features can be clearly represented and accurately extracted is the basis of fingerprint recognition technology [1]. This is because the accuracy of the fingerprint's feature points has a great deal to do with the results of fingerprint image processing.

The purpose of fingerprint image denoising is to remove all kinds of noise pollution, and to process the original image into a black-and-white binary map with obvious feature points as far as possible, so that the feature points can be accurately extracted later. In the denoising process, both noise cancellation is ensured and the true character of the original image is not compromised as much as possible. Moreover, whether the noise removal effect is good or bad, and whether it is accurate or not, depends on the discrimination rate and final result of the fingerprint recognition system.

Li et al.[2] and Huang et al.[3] proposed new wavelet threshold shrinkage functions based on wavelet threshold denoising theory, respectively, which improved the denoising effect of fingerprint images to a certain extent. The method of fingerprint denoising based on the ridge direction clustering dictionary is proposed, which has a higher peak signal-to-noise ratio than the traditional threshold function method.Although these methods are more robust, they have difficulty recovering the detailed features of the fingerprint image while removing noise.

With deeplearning, especially convolutional neural networks (CNNs) have achieved good results in the field of image recognition, many people have also applied CNNs in fingerprint denoising processing.The results of this study are as follows: Jain et al.[6] proposed the problem of denoising natural images using convolutional neural networks, looking at the denoising process as a neural network fitting process to improve the signal-to-noise ratio of images;Li etal.[7] proposed a deep convolutional codec network for image denoising, using a combination of sparse coding and self-encoding to achieve image denoising;Xu etal[8] proposed to use deeper convolutional neural networks for denoising, with better PSNR values and visual effects compared to other current methods;Lore et al.[9] used a autoencoder method of deeplearning to train features of different low-illumination image signals for adaptive brightening and denoising.Although these methods employ deeplearning and use more complex models, the improvement in noise removal performance is limited.

Therefore, inspired by the application of convolutional neural networks in image denoising, In this paper, we have designs and proposes a fingerprint denoising algorithm based on deep encoder-decoder neural network. The encoder subnet and decoder subnet form a symmetric deep convolutional neural network structure to learn the nonlinear mapping relationship between the fingerprint image and its noisy image.

# 1 RELATED WORK

## 1.1 Difficulties of fingerprint image denoising

The conventional fingerprint image denoising algorithm is roughly divided into two steps: first, select the appropriate number of filter cores and decomposition layers according to the noise characteristics, and then filter the noise-containing image to get the filter weight factor. Then the high-frequency coefficients of the decomposed layers are thresholded to obtain new filter coefficients, and the image is filtered and reconstructed according to the low-frequency coefficients of the filter decomposition and the high-frequency coefficients after the threshold processing to achieve image denoising. However, in the actual scene, the fingerprint collection will be affected by dirty fingers, scars on the fingers, too dry or wet fingers and the environment and other factors, the collected fingerprint image will be disturbed by different degrees of noise, and these traditional image denoising methods often use a fixed filtering weight factor or human experience to set the filtering core to process the image of fixed noise features, so it is not easy to handle non-fixed noise features. Unlike other natural images, fingerprint images are usually composed of fine textures and edges, and during the denoising process, the fingerprint features of the original image should be kept as undamaged as possible so that the fingerprint recognition system can recognize the fingerprint based on these features, which makes denoising more difficult.

Because of the above-mentioned problems with fingerprint noise, it is difficult to design effective filter cores, resulting in poor performance of conventional denoising algorithms.

## 1.2 Convolutional neural networks

Convolutional neural network[10] is an efficient identification method that has been developed in recent years and has attracted widespread attention, especially in the field of image processing, where it is more widely used because it avoids complex pre-processing of the image and can be directly imported into the original image for processing. Figure 1 shows a common convolutional neural network structure, where the input of each neuron in the network is connected to the local receptor field of the previous layer, and several feature maps are generated by extracting image features through convolutional operations. convolutional neural networks enable the learning and characterization of image features through a combination of convolutional, pooled and fully connected layers.

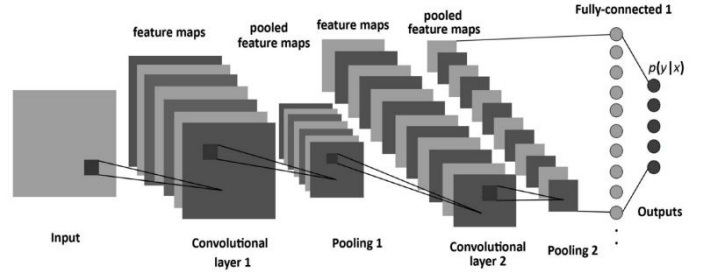

**Fig. 1** convolution neural network

## 1.3 Residual Learning

The proposal of convolutional neural networks has triggered a series of breakthroughs in image processing methods, which have more powerful feature learning and feature expression capabilities than traditional image processing methods. However, as the network layers deepen, there is a decrease in accuracy, and deep network training becomes very difficult[11]. Therefore, to address this issue, He et al[12] proposed a residual learning framework with the structure shown in Figure 2. Assuming that the input of this part of the neural network is x, the function mapping (i.e., the output) H(x) to be fitted can be defined as another residual mapping F(x) as H(x)-x, and the original function mapping H(x) can be expressed as F(x)+x. And it is much easier to optimize the residual mapping F(x) than the original mapping H(x). F(x)+x can be understood as the sum of shortcut x and main path F(x) in a feedforward neural network. The shortcut does not introduce additional parameters, does not affect the complexity of the original network, and the overall network can still be solved using the existing deeplearning feedback training.So residual learning by adding shortcut connections to the original convolutional layer to form the basic residual block, and then by accumulating the residual block structure layer by layer in the network, not only effectively avoid the problem of gradient explosion or disappearance caused by too many layers, improve network performance and speed up training, but also make better use of low-dimensional image features and improve accuracy.

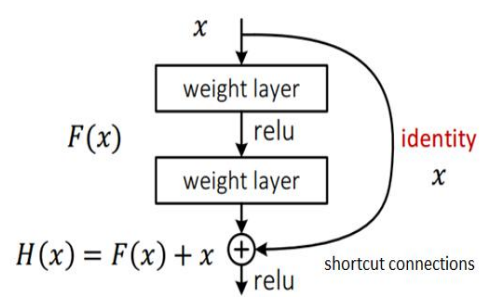

**Fig. 2** residual structure

### 1.4 Dilated Convolution

The dilated convolution be used in the image segmentation, and there are 2 key steps in image segmentation: pooling operation to increase the field and upsampling operation to increase the image size.Although the image was up-sampled to restore size, there are still many detailed features that were lost due to the pooling operation,To solve this problem, Yu et al.[13] proposed a method of dilated convolution that adds a coefficient of expansion to the original convolution compared to the ordinary convolution, which can expand the convolution nucleus to the scale bounded by the coefficient of expansion and fill the unoccupied area of the original convolution with 0,In this way, the dilated convolution has a larger perceptual field than the ordinary convolution, and by setting different coefficients of expansion, it is possible to extract multi-scale contextual information of the image and make better use of the image information, as shown in Figure 3.

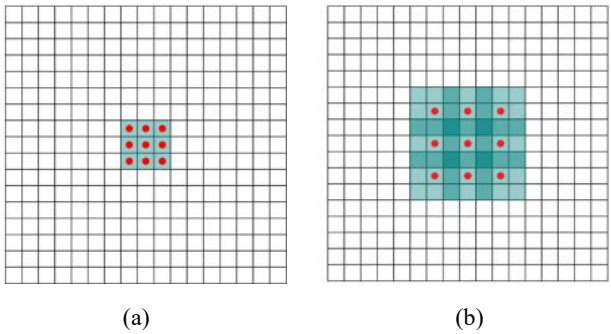

(a)　　　　　　　　　(b)

**Fig. 3**　dilated convolution

The visual field effect of the classical 3×3 convolutional core role is shown in Figure 1(a), each time the 3×3 visual field is covered; the corresponding 3×3 expansion coefficient in Figure 1(b) is a 2 times dilation rate convolution, in fact the size of the convolutional core is still 3×3, but the calculated visual field of the convolutional core increases to 7×7, while the actual parameter is only 3×3.In addition, it is easy to see that the simultaneous effect of a convolution with a coefficient of expansion of 1 and a coefficient of expansion of 2 is the same as the effect of a conventional convolutional core acting 7×7 alone, so that the characteristics of different fields of view can be felt by setting convolutional cores with different coefficients of expansion acting on the same input.

## 2　Proposed Method

In this paper, we take full advantage of the convolutional neural network and the above methods to design and propose a fingerprint image denoising network based deep convolutional neural network, which consists of a encoder subnet and a decoder subnet, and the network structure is shown in Figure 4.

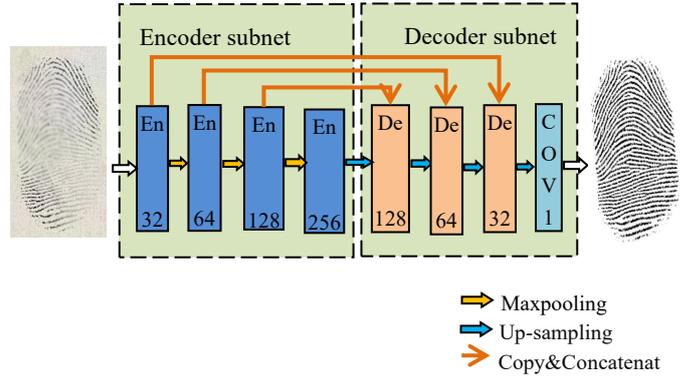

**Fig. 4**　An illustration of the proposed framework

### 2.1　Encoder subnet

The Encoder subnet consists of four encoder blocks as shown in Figure 5, and each encoder blocks consists of three convolutional layers with three different times dilation rate, a BNom layer and a PRELU layer, and a Dropout layer after each convolutional layer to improve the robustness of the network and prevent overfitting. Different times dilation rate are set for different convolution layers, with a convolution of 1 times dilation rate to extract the detailed texture features of the fingerprint, a convolution of 2 and 5 times dilation rate to extract the texture and edge features of the fingerprint, and a combination of convolutions of different coefficients of expansion to capture multi-scale contextual information of the image. The encoder blocks are connected by max-pooling pooling operations, with the pooling window size set to 2×2. As the number of encoder blocks increases, the encoder blocks ability to characterize the image becomes stronger, doubling the convolutional core channel with each pooling operation until it equals 256.

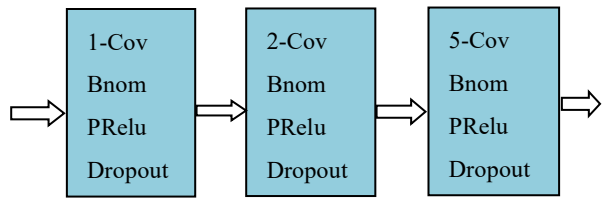

**Fig. 5**　encoder block

### 2.2　Decoder subnet

The Decoder subnet up-samples the features extracted by the encoder subnet by performing an inverted convolution operation, and the higher-order features extracted by the decoder subnet are mapped layer by layer into images of the same size. The decoder subnet consists of three decoder blocks, and the blocks are connected by up-sampling operation. In the deconvolution process, to make better use of the low-order spatial information of the image to guide the process of recovering the detailed features of the fingerprint, residual learning is introduced in the network. First, in each decoder block, the information of the previous decoder block is fused with the information of the next decoder block by means of residual connection, and second, in the whole network, the information of the encoder block is fused with the corresponding decoder block by means of constructing a replication channel, so that the shallow contextual information can spread to higher resolution layers, and the deep abstract features are combined with shallow features. The last convolutional layer with convolutional kernel number of 1 outputs the image using the sigmoid activation function to achieve optimal denoising by comparing the difference between the output of the network and the original noiseless image.

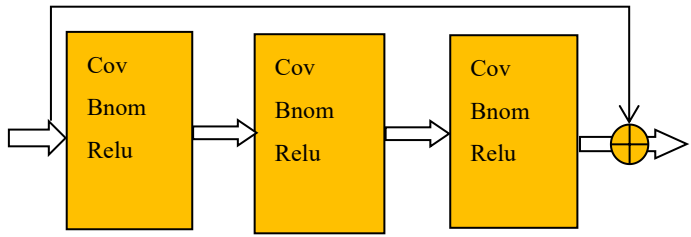

**Fig. 6** decoder block

## 3 EXPERIMENTSANDDISCUSSIONS

### 3.1 Dataset Description

The dataset in we adopts the large-scale generated realistic artificial fingerprint dataset provided by CONDALAB[14], which consists of 92400 original fingerprint images and their corresponding noisy fingerprint images generated in various complex backgrounds to simulate the fingerprint images collected under real scenes (e.g., walls, skin), as shown in Figure 7.

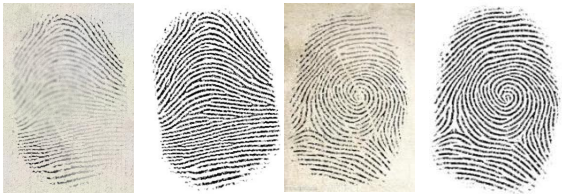

**Fig. 7** data example

### 3.2 Evaluation Metrics

To fully verify the superiority of we proposed deep encoder-decoder neural network, we used two evaluation metrics, peak signal-to-noise ratio (PSNR) and structural similarity (SSIM), as well as the comparison of the visual effects after denoising, were used for evaluation and analysis.The larger the two indicators, PSNR and SSIM, the better the noise removal effect of the algorithm,PSNR is defined as the ratio between the peak rate of the image and the noise rate in the image:

$$\text{PSNR} = 10 \sum_k \lg \left[ \frac{\max DR^2}{\frac{1}{m} \sum_{i,j} (t_{i,j,k} - y_{i,j,k})^2} \right] \quad (1)$$

SSIM is defined as a multiplicative combination of similarities in contrast, brightness and structure between an image and a reference image:

$$\text{SSIM}(x,y) = [l_M(x,y)]\alpha_M \bullet \prod_{j=1}^{M}[c_j(x,y)]\beta_j[s_j(x,y)]\gamma_j \quad (2)$$

where DR is the dynamic range, m is the total number of pixels in each of the three color channels, $l_M(x,y)$ is comparison on the brightness component, $c_j(x,y)$ is comparison on the contrast component, $s_j(x,y)$ is comparison on the structure component, and $\beta_j, \gamma_j$ are adjustment factor, respectively.

### 3.3 Parameter Settings

The experimental computer is Intel i7 CPU, NVIDIA GeForce 1060 GPU, 8GB RAM. The network is written using the Keras advanced neural network API interface and implemented as a backend using the TensorFlow deeplearning framework, with parallel accelerated calculations using CUDA and CUDNN during training and testing.

During the experiment, 80% of the image data of the dataset is randomly selected as a training set for training the network model, 10% as a validation set for testing the current performance of the network in the training phase to monitor the training process of the network and decide whether to terminate the training, and 10% as a test set for the final performance evaluation of the training results, while data enhancement processing is performed on the training data to expand the training sample through a series of rotation, scaling, flipping, etc.

For network loss selection, we uses the mean square error (MSE) for network training, which is defined as in equation (3):

$$MSE = \frac{1}{MN} \sum_{i=1}^{M} \sum_{j=1}^{N} \left( y_t - \hat{y}_t \right)^2 \quad (3)$$

where M×N is the size of the image, $\hat{y}_t$ is predicted value of the network for each pixel, and $y_t$ is actual value of each pixel.

For the choice of loss optimization algorithm, we uses Adam algorithm with the initial learning rate set to 1e-3 and the learning rate decreasing to half of the original every 3 rounds as the number of training rounds changes.The image size set to 256x256,batch size is set to 8, the random deactivation ratio is 0.3, and in addition, to prevent overfitting, the EarlyStopping mechanism in keras is introduced to stop training when the training set losses no longer decreases at 5 consecutive iterations of the loss value. Also to save the optimal network parameters, the ModelCheckpoint mechanism in keras is introduced, and after each iteration is completed, it is decided whether to save the current network parameters by monitoring whether the losses on the verification set drop or not.

### 3.4 Comparison of Results

In order to verify the effectiveness and superiority of our proposed algorithm in fingerprint image denoising, a comparison experiment was conducted between U-Finger [15], CVxTZ [16], Krishnapriya [17] and our proposed algorithm.

Table 1　Comparison of denoising performance of each method

| Method | U-Finger | CVxTZ | Krishnapriya | Our work |
|---|---|---|---|---|
| PSNR | 16.9768 | 16.6094 | 16.6068 | **17.3594** |
| SSIM | 0.8266 | 0.8093 | 0.8027 | **0.8319** |

From the comparison results in Table 1, it can be seen that the PSNR of the proposed method can reach 17.6594, which is 0.6826 higher than the best performance of the other three methods, and the SSIM of the proposed method is 0.8419, which is 0.0153 higher than the best performance of the other three methods.The main reasons are twofold: first, the deep convolutional neural network can identify complex picture features and fully extract the deep fingerprint features of noisy images, and the dilated convolution increases the receptor field without increasing the complexity, effectively taking into account the global and detail features of the fingerprint extraction; second, by introducing residual learning, the features of different levels of the network can be fused, thus further recovering the detail features of the fingerprint and improving the noise removal effect.

### 3.5 Visual Comparison

To visualize the denoising effect of our algorithm, the test set image data is denoised using our algorithm and the comparison algorithm, and the denoising results of some test images are shown in Figure 9.

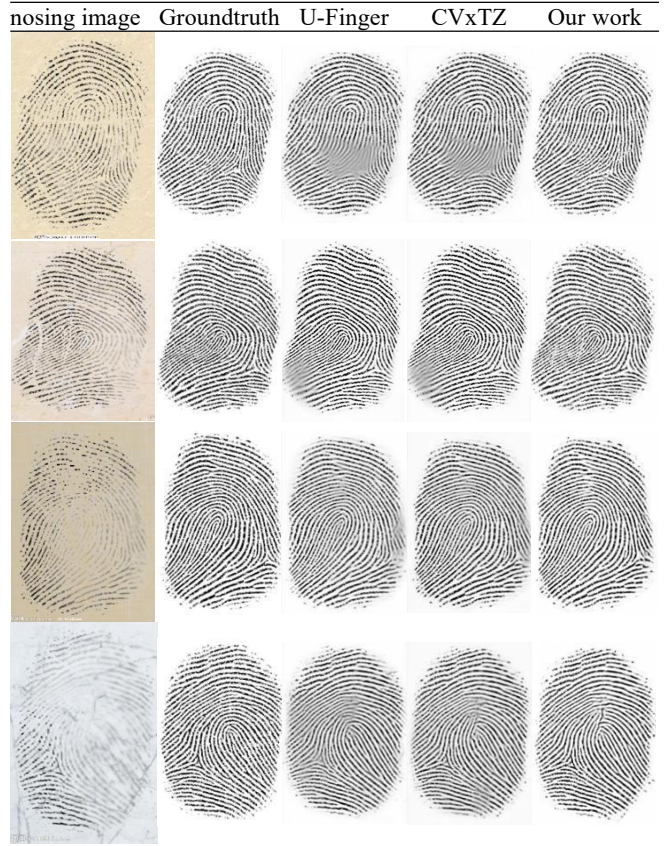

**Fig. 8**　comparison of denoising effect

As can be seen from the comparison results in Figure 8, the CVxTZ and U-Finger fingerprint denoising methods, while removing most of the noise and retaining fingerprint texture information and edge information, are prone to background artifacts.we proposed method effectively obtains the global field of view of the noisy picture by means of void convolution and residual learning for the network, and realizes the field of view information of different sizes through the combination of void convolution of different sizes, which effectively solves the aforementioned defects. Both the edge and grain details of the fingerprint achieve good noise removal, better restoring the characteristic information such as edges, lines and curves of the fingerprint texture.

## 4　CONCLUSION

In order to improve the denoising effect of fingerprint image, we proposes a symmetric deep Encoder-Decoder fingerprint denoising network consisting of encoder subnet and decoder subnets; the encoder subnet are used to extract the deep fingerprint features of the noisy image, and the decoder subnet reconstruct the original fingerprint image based on these deep fingerprint feature images to achieve fingerprint image denoising; at the same time, dilated convolution and residual learning are introduced into the network to increase the receptor field without increasing the complexity, so as to achieve feature fusion at different layer of the network to further recover the detailed features of the fingerprint and improve the denoising effect. The experimental results show that the PSNR and SSIM can achieve 17.6594 and 0.8419, while the edges, lines and curves in the fingerprint image can be better recovered, with better visual effects compared to other methods.